\documentclass[conference]{IEEEtran}
\IEEEoverridecommandlockouts
\usepackage{cite}
\usepackage{amsmath,amssymb,amsfonts}
\usepackage{algorithmic}
\usepackage[numbers]{natbib}
\usepackage{url}

\usepackage{graphicx}
\usepackage{textcomp}
\usepackage{xcolor}
\def\BibTeX{{\rm B\kern-.05em{\sc i\kern-.025em b}\kern-.08em
    T\kern-.1667em\lower.7ex\hbox{E}\kern-.125emX}}
\begin{document}

\title{Manipulating a Tetris-Inspired 3D Video Representation \\

}

\author{\IEEEauthorblockN{Mihir Godbole}
\IEEEauthorblockA{\textit{ Department of Computer Science \& Engineering} \\
\textit{Texas A\&M University}\\
College Station, TX, USA \\
amigo2000@tamu.edu}}

\maketitle

\begin{abstract}
Video Synopsis is a technique that performs video compression in a way that preserves the activity in the video. This technique is particularly useful in surveillance and monitoring applications. Although it is still a nascent field of research, there have been several approaches proposed over the last two decades varying with the application, optimization type, nature of data feed, etc. The primary data required for these algorithms arises from some sort of object tracking method. In this paper, we discuss different spatio-temporal data representations suitable for different applications. We also present a formal definition for the video synopsis algorithm. We further discuss the assumptions and modifications to this definition required for a simpler version of the problem. 
We explore the application of a packing algorithm to solve the problem of video synopsis. Since the nature of the data is three dimensional, we consider 3D packing problems in the discussion. This paper also provides an extensive literature review of different video synopsis methods and packing problems. Lastly, we examine the different applications of this algorithm and how the different data representations discussed earlier can make the problem simpler. We also discuss the future directions of research that can be explored following this discussion.

\end{abstract}


\section{Introduction}
In the modern era, the rate of data generation through different sources is perhaps the highest as compared to previous time periods. Not only has the sheer quantity of data increased, but also its modalities. One such modality is the visual modality which comprises image and video data. The generation of this form of data is generally through the use of cameras to either capture photos or record video footage. There are other sources of generation as well such as computer generated statistical images or diagrams, medical scans and digital documents. It is crucial to analyze and understand the underlying patterns in this data to develop solutions that will aid mankind. Analyzing image data involves a range of techniques such as object detection, image segmentation, image pre-processing, image enhancement, etc. Video data on the other hand is used mostly for object tracking and trajectory analysis. Manual monitoring and analysis of video data is an error prone and tedious task. To tackle this problem, the idea of Video Synopsis has emerged where a video is condensed to a smaller duration video while preserving every object activity in the video. The two primary steps in any Video Synopsis system, data representation and the synopsis algorithm will be discussed in this paper. Along with that, the different applications resulting from this approach will be investigated.

The data required for the synopsis algorithm is not in the form of images or frames, but it is a form of sequential numerical data which represents the different object trajectories in the video. These trajectories are referred to as ‘tubes’\cite{4408934}. It is assumed that we already have the data that represents the object trajectories. In practice this data is generated using object tracking algorithms, which do not always yield continuous tubes. Moreover, there are challenges such as object occlusion which affect the object tracking algorithm as this may lead to incomplete tracks. The requirement of any synopsis algorithm is clean and  complete tubes, which itself presents a research problem. However, the scope of this paper is limited to the data representation and synopsis algorithm segments of the Video Synopsis pipeline. The representation of these tubes in a topological space will depend on the application of the solution. In real world video footage, the objects may not move in a pattern that will generate “nice” geometrical shapes for the tubes. In this paper the extent of approximation in the shape of such tubes that can work for the synopsis algorithm will be discussed. As a part of the approximations, the thickness of the tubes which correspond to the object's dimensions in the 2D place can also be varied. It is important to consider the effect of these variations on the cost associated with the synopsis algorithm. While in most applications the configuration space for these objects can be left untouched for an accurate representation of the tubes, this paper will explore applications that will need some identification on the configuration space to satisfy the requirements of accurate data representation.

The crux of a Video Synopsis system is the synopsis algorithm. This problem can be viewed as a complex Tetris game where the aim is to stack the objects as compactly as possible. A synopsis algorithm also aims to ‘stack’ the tubes as compactly as possible. This problem can also be interpreted as a scheduling, energy minimization or even a packing problem. Depending on the application, the constraints for this problem may vary, but most approaches have presented a solution with respect to the constraints of time and the number of collisions. Unlike tetris, we have irregular shaped 3D object tubes to stack. Taking inspiration from approaches such as \cite{ZHAO2021101234}, this paper will present a solution to the problem of video synopsis using an approach modeled on 3D packing. 

With the pace of the world increasing everyday, reducing the time required to complete day-to-day tasks and jobs is important. From jobs ranging from security surveillance to satellite feed monitoring and analysis, a tool like Video Synopsis will largely reduce the time and efforts required for manual monitoring. This paper will discuss the applications of this approach in different areas ranging from sports to traffic management and the modifications required for every application.


\section{Related Works}

\subsection{Video Synopsis}
Video synopsis can be understood as a well defined,
multi-stage pipeline. Beginning with the raw video as input,
the first stage is object detection followed by object tracking.
Object detection is one of the primary applications of computer
vision. Object tracking mainly uses heuristic algorithms, with recent approaches using deep learning methods as well. These
two stages are usually combined and are done in real-time. The
data from these two stages represents the ‘activity’ in the video.
This data is used in the optimization stage which constitutes a synopsis algorithm.

Optimization forms the core of the video synopsis generation pipeline. Collisions between the rearranged tubes or object tracks pose the biggest challenge. Although video condensation techniques include both frame-based as well as object-based methods, we will consider only object-based methods as our focus is activity or object-level optimization. Frame-based optimization methods such as \cite{truong2007video, nam1999video} aim to summarize the video by selecting the key frames from the video which capture the necessary activity. This does not preserve all the activity in a video.  Rav-Acha et al. \cite{ravacha_2006_making} were among the first to propose a novel method of dynamic video synopsis via energy minimization.

\citet{yogameena_2020_abnormal} propose an activity-based video synopsis technique that makes use of Seam Carving for pixel-based analysis on a real-time dataset without temporal shifting of objects. The algorithm is primarily applicable to abnormal activity detection.  

\citet{yang_2021_scene} suggest global and dynamic rearrangement of tubes on collision detection using a three-dimensional Octree, which is a hierarchical data structure involving division of space using recursion. Minimal memory is required for representing spatio-temporal information of entire tubes as a cuboid in the Octree. Rearranging tubes dynamically makes more efficient use of synopsis space compared to tubes with significant spatio-temporal realignment for collision avoidance that is known to increase the complexity and the length of the condensed video.

\citet{ra_2018_parallelized} suggest an online tube rearrangement strategy based on Fast Fourier Transform (FFT). While the presented performance metrics like Frame Condensation Ratio, Compact Ratio, and Overlap Ratio were comparable to the other approaches, iterative optimization caused discontinuity of motion flow when small sections of the condensed video were merged into a larger whole.

An online synopsis algorithm based on dynamic graph
coloring is proposed by  \citet{ruan_2019_rearranging}. The dynamically generated graph consists of object masks of tubes represented as vertices(V) and relationships between them preserved as edges(E). Video synopsis in this case can be seen as a step-
wise optimization problem on a graph sequence seeking to achieve minimum collision in the spatial domain and maximum condensation rate. This method, however, performs worse than the online baselines considered by the authors.

\citet{jianqingzhu_2015_highperformance} present an online optimization technique for high performance video condensation using multi-threading to overcome drawbacks of offline processing such as memory load and slow processing speed. The duration of the final video is set automatically, like in the work done by \citet{li_2018_video}. Tube rearrangement is performed using a tube filling algorithm inspired by the Tetris game where tubes are optimally arranged to make efficient use of the three dimensional condensation space using a greedy strategy.

A video synopsis technique that incorporates the speed
and size of the object during rearrangement is used by \citet{nie_2020_collisionfree}. An object is shifted, its speed altered and size rescaled simultaneously to reduce collisions between trajectories (extracted tubes) and disruption of chronology with respect to the other entities. The Metropolis sampling algorithm, based on the Markov chain Monte Carlo (MCMC) method, is used for the optimization of a unified energy function that takes into consideration all three variables to achieve a high compression ratio.

\citet{ghatak_2021_gan} further propose another optimization approach for object-based offline synopsis. Hybridization of Simulated Annealing and Grey Wolf Optimizer (HGWOSA) is applied to ensure faster convergence at global optimum with low computational cost by making use of the advantages offered by both algorithms. The aim is to minimize the cost of temporal consistency and collision while preserving all activities from the original video.

In the approach proposed by \cite{moussa_2020_objectbased}, objects are rearranged using Particle Swarm Optimization as it performs better than other heuristic algorithms such as GA in terms of speed of convergence at global optimum and computational overhead. The proposed approach aims at achieving an optimal trade-off between the chronological order of tubes and collisions by minimizing a variant of the traditional energy function based on temporal consistency, collisions, synopsis length, and true collision costs to preserve original interactions to the best possible extent.

While most traditional approaches focus on general video
condensation, \citet{ahmed_2020_querybased} and \citet{namitha_2021_interactive}
suggest query-based techniques that generate targeted synopsis
videos based on user requirements for filtered optimization to
reduce computational overhead and stitch only those objects
that are requested by the user based on characteristic data,
time, or position. Hence, only required activity is preserved
as compared to conventional approaches that seek to preserve
most activity from the original video.

The aforementioned techniques are designed for a single camera footage. Methods such as \cite{zhu_2016_multicamera, zhang_2020_multiview, singhparihar_2021_multiview, hussain_2020_intelligent, dilber_2021_a} propose synopsis algorithms for multiple cameras, but the scope of this paper is limited to single camera or single view video synopsis. 



\subsection{Packing Algorithms}

Packing algorithms belong to a category of algorithms that are utilized in optimization challenges. These challenges involve packing items of varying sizes into a limited number of bins or containers. Each bin or container has a predetermined capacity. The goal of these algorithms is to minimize the number of bins used for packing. In the case of video synopsis, we will consider a single bin to pack the objects. Packing problems can be categorized into two types: 2D packing problems, and 3D packing problems. 

\subsubsection{2D Packing Problem}
\begin{figure}[htbp]
\centerline{\includegraphics[width=0.3\textwidth]{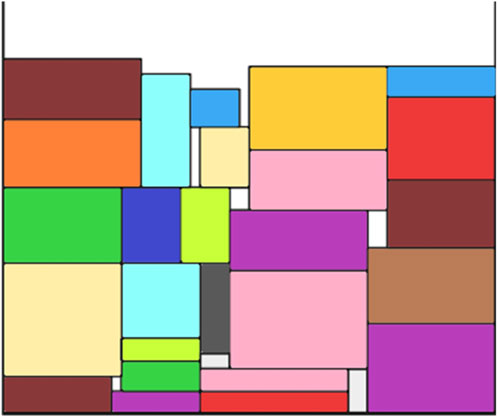}}
\caption{2D regular sized packing problem \cite{guo20222d}}
\label{fig1}
\end{figure}

\begin{figure}[htbp]
\centerline{\includegraphics[width=0.4\textwidth]{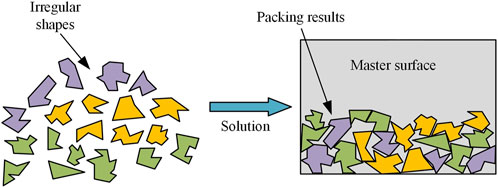}}
\caption{2D irregular shape packing problem \cite{guo20222d}}
\label{fig2}
\end{figure}

This classification can be further dissected into regular and irregular shaped packing problems. A representation of each is shown in Fig.~\ref{fig1} and Fig.~\ref{fig2}. The 2D packing problem algorithms are particularly useful in the industrial and manufacturing sector. Packing rectangular polygons has been explored by \cite{wu2016novel, wu2017improved, martello2015models}. \citet{wu2017improved} improves on the previous work \cite{wu2016novel} by introducing a three-rectangle placement mechanism for 2D rectangle packing area minimization problems with central rectangles (CR-RPAMP). Regular shaped packing problems also include layout problems for circles \cite{zhang2004packing, nurmela1997packing, fekete2019split}. 2D packing problem is mentioned to be a NP-hard problem  in \cite{guo20222d}, although the problem of packing circles in a square was shown to be NP-hard by \cite{demaine2010circle}. Similarly, the decision problem whether it is possible to pack a given set of squares into the unit square was shown to be strongly NP-complete by \cite{leung1990packing}.

The problem of two-dimensional irregular packing has been a longstanding focus in mathematical and combinatorial optimization, evolving over decades. Initial approaches to solving the 2D irregular packing problem involved the application of a single algorithm, such as linear programming \cite{gomes2006solving, cherri2018mixed}, a meta-heuristic like Genetic Algorithm \cite{pinyapod2014efficient, bortfeldt2006genetic}, or a heuristic algorithm. One of the simplest methods is the envelope polygon method \cite{peralta2018solving, adamowicz1976nesting}, which converts the problem to a regular shape packing problem by enveloping the irregular shape with a regular shape. \citet{math11020327} propose a hybrid reinforcement learning algorithm, creating a foundation for future deep reinforcement learning based solutions to 2D packing problems.

\subsubsection{3D Packing Problem}

3D packing problem is considered to be NP-hard \cite{liu20233dbin}.  Similar to 2D packing, the algorithms can be categorized based on the shape of the objects to be packed as regular or irregular. A 3D irregular object is one that requires more than three variables to describe its shape. 
\citet{doi:10.1080/00207543.2018.1534016} provide a literature review of the 3D irregular packing methods for additive manufacturing. Similar to 2D packing, there are solutions proposed using meta-heuristic \cite{ZHAO2021101234, lamas2023voxel} and heuristic \cite{XiaoLIU:380} algorithms. \citet{edelkamp2015packing} approximate shapes into a sphere similar to the envelope polygon method used for 2D packing problems. In terms of 3D object representation, many approaches have used a mesh, voxel, or level set representations. The voxel structure has been widely used \cite{wu2014vox,lamas2023voxel}. \citet{ZHAO2021101234} have used a mesh representation processed from 3D imaging scans.

It is important to note the similarities between the objective of the synopsis algorithm and 3D irregular packing problems. The minimization of height in 3D packing is analogous to the minimization of time length in video synopsis. Some solutions allow free rotation for the objects, which is in most cases not allowed in video synopsis. Therefore some constraints will have to be added to these solutions and the underlying algorithms when translating them for the case of video synopsis.

\section{Problem Description}

\subsection{Data Representation}\label{AA}

The video data is a collection of 2D frames captured by a camera. The duration of the video or the sequence length is the third dimension that will be used to represent the time dimension. Thus objects moving in the video are represented in a 3 dimensional space with $x, y$ and $t$ as the axes. In this section, we will discuss the variations in this data seen in different applications as well as the different shapes we get depending on the choice of object detection and tracking algorithms. 

\subsubsection{Input data}
\begin{figure}[htbp]
\centerline{\includegraphics[width=0.4\textwidth]{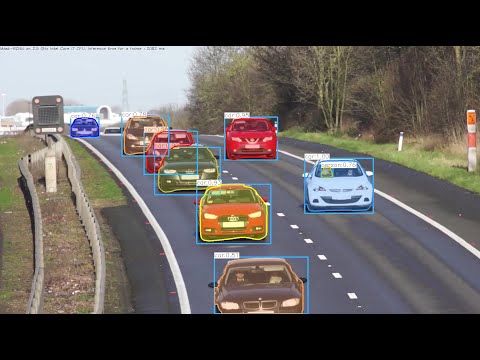}}
\caption{Object Detection Masks and Bounding Boxes}
\label{fig3}
\end{figure}

We will first discuss the nature of the data we get from the object detection and tracking algorithms. The detected objects are marked either by a bounding box or a segmentation mask as shown in Fig.~\ref{fig3}.
\begin{figure}[htbp]
\centerline{\includegraphics[width=0.5\textwidth]{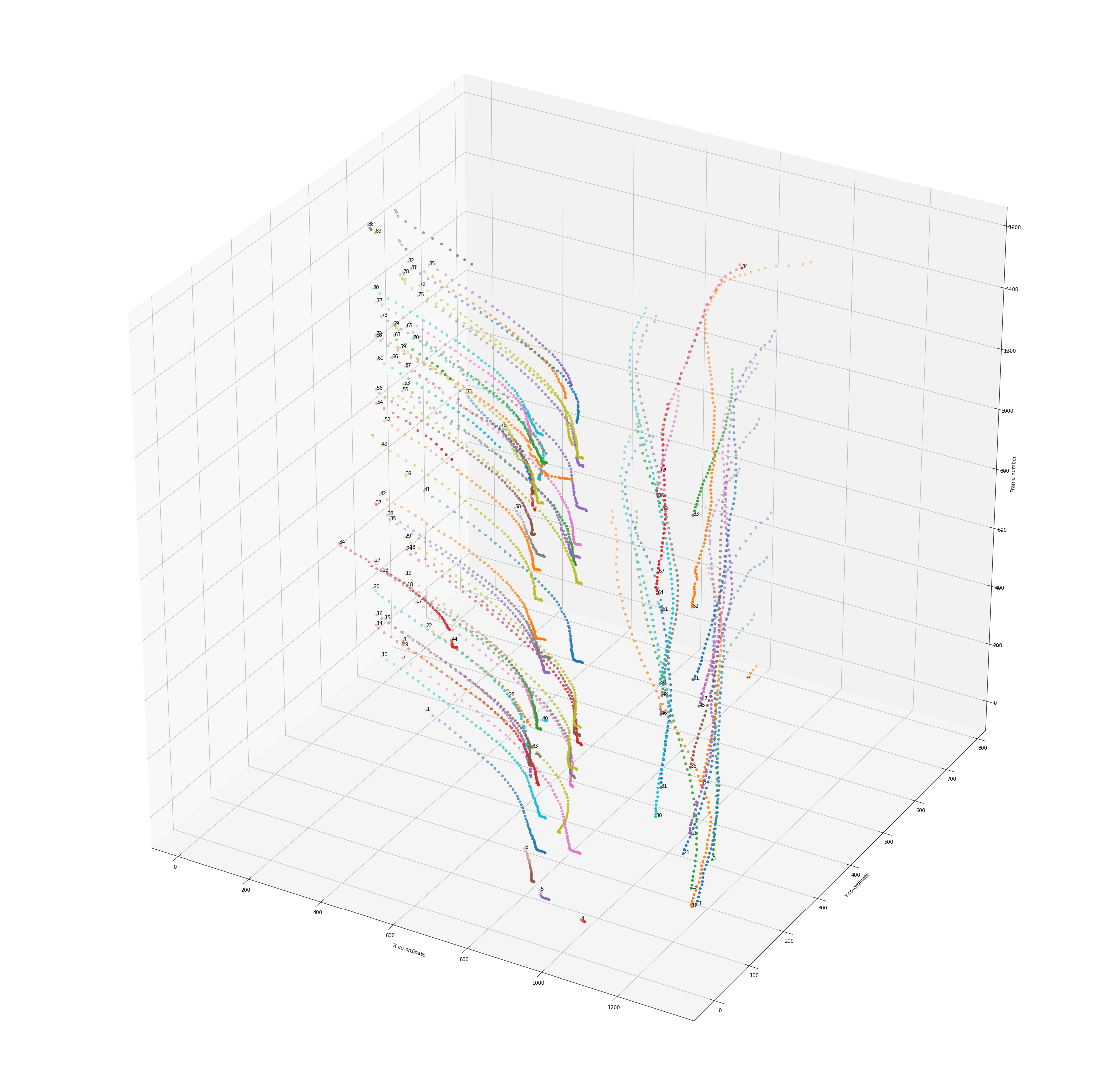}}
\caption{Centroid tracks}
\label{fig4}
\end{figure}
In the simplest case, we can represent the object tracks using the centroids obtained from either the bounding boxes or the segmentation masks. Centroid tracks that represent object activity in highway footage are illustrated in Fig.~\ref{fig4}.
Although this gives a clear sense of trajectories, this reduces the quality of the resulting video because of the large overlap between objects. Collisions will be formalized in the next section.  Next, we will consider the bounding boxes which simply draw a rectangular box around the extremities of an object. It is important to note that most object detection algorithms like YOLO \cite{Redmon2015YouOL}, and MaskRCNN \cite{8237584} produce Axis-Aligned Bounding Boxes (AABB). This means that no matter how the object is rotated in the 2D plane, its bounding box will always be parallel to the axes of the plane. Some works \cite{s18082702} have used rotated bounding boxes for their particular applications. 
The key difference between the two cases is that as the object in the video rotates in the 2D plane, the bounding box axes remain parallel to the frame axes in the first case. The bounding box is transformed into a rectangle of new dimensions as the object rotates. On the other hand, in the second case, the object track looks like a twisted tube with the box rotating with the object. \\
\subsubsection{Special Cases}
\begin{figure}[htbp]
\centerline{\includegraphics[width=0.5\textwidth]{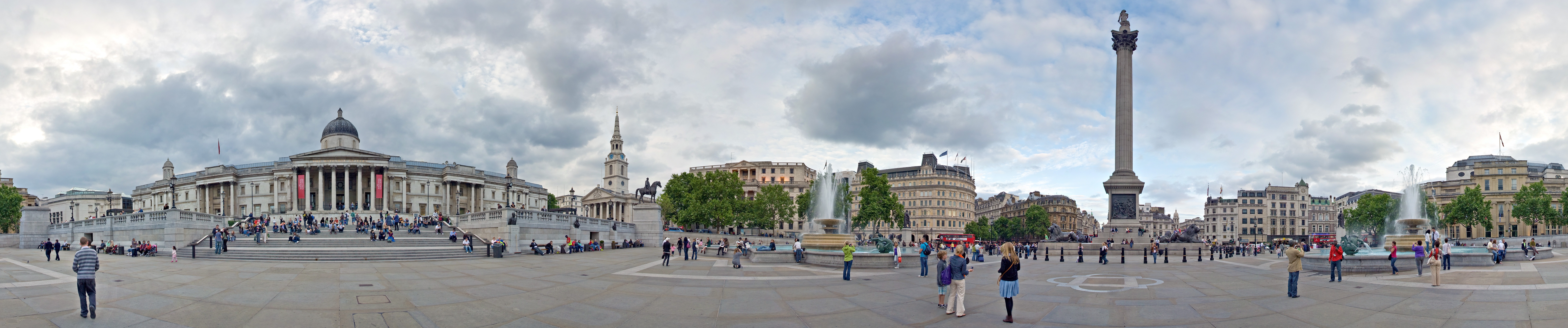}}
\caption{Panaroma image}
\label{fig6}
\end{figure}

We will consider some special cases which might require a different data representation. So far we have discussed cases where the field of view is limited to an angle less than 180 degrees. Panoramic or 360 degrees cameras capture a circular or spherical view of the scene. We will consider the case of panoramic cameras. Although we have a circular 360 degrees view from the camera, it will still be represented in a 2D frame as shown in Fig.~\ref{fig6}.

\begin{figure}[htbp]
\centerline{\includegraphics[width=0.4\textwidth]{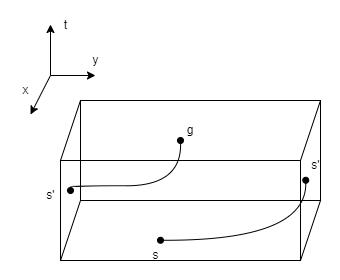}}
\caption{Representations of activity from a panoramic camera}
\label{fig7}
\end{figure}

The activity in such a frame can be represented in the 3D space we have discussed above. As can be seen in Fig.~\ref{fig7}, An activity representing the same object gets split into two tracks as the object exits from one edge and enters from the other.

\begin{figure}[htbp]
\centerline{\includegraphics[width=0.3\textwidth]{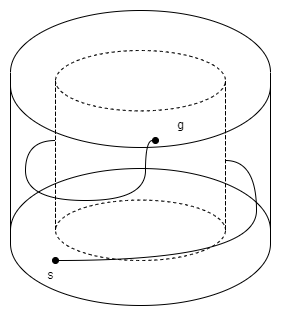}}
\caption{A rectangular torus representation of the configuration space for object tracks}
\label{fig8}
\end{figure}

This can be resolved by the identification on the panoramic image to get a cylinder. Identification is a way to redefine a topological space by saying that some points in a space are identical \cite{lavalle:2006}. An identification on the 3D space representing the data will give us a rectangular torus. Thus the activity can be represented as one continuous track. This is demonstrated in Fig.~\ref{fig8}. 

A similar transformation of the configuration space will be required for any application where the frame has "no boundaries". An example of this is the Pac-Man game.

\subsection{Synopsis Algorithm}
A simple way to understand the synopsis algorithm is to think of the Tetris game. In a Tetris game, the goal is to fit the new blocks as compactly as possible to reduce the total height of the stack. Similarly, in the case of video synopsis, we have a three dimensional tetris. The Tetris game can be thought of as a real-time synopsis problem as the blocks are revealed sequentially. Video Synopsis can be offline or online (real-time). Therefore another way to think about it is you are given $n$ boxes of different shapes and sizes and you need to pack them efficiently to reduce the stack height. This thus becomes a packing problem. The only difference is that here the boxes are constrained in the spatial axes. They cannot be moved in in the $x$ or $y$ directions as you can in a tetris game or a packing problem. Therefore it is a special case of both. Before discussing the proposed algorithm, the different constraints must be discussed.
\\
The two major constrains are time and the number of collisions,
\subsubsection{Time}
The primary goal of this algorithm is to reduce the time length required to represent all the activity or tubes. In the best case scenario, it will be the maximum time length of a single tube. In the case where the tubes have a length comparable to that of the full length video, speeding up or resizing/shrinking the tubes in the time dimension can be considered. We can therefore define two constraints:
\begin{itemize}
    \item The maximum time length of the synopsis video.
    \item The compression factor for long tubes.
\end{itemize}

\subsubsection{Collisions}
If two tubes overlap each other at any time step $t$ with an overlap area greater than a predefined permissible overlap area, then the tubes are said to be in collision with each other. Thus we have two constraints:
\begin{itemize}
    \item The maximum number of collisions allowed.
    \item The maximum permissible tube overlap.
\end{itemize}
Other than the above constraints, some other factors that should be considered before designing the synopsis algorithm are:
\subsubsection{Order of the tubes} If the chronological order in which the objects appeared in the original video is to be maintained, a new constraint has to be added to every tube specifying the rank/order in which the objects must appear. 
\subsubsection{Real-time vs Offline} 
In the real-time scenario, we have to pack the tubes one by one with knowledge of only a few subsequent tubes. In the offline case, we have all the geometric representations of all the tubes beforehand.

\section{Problem Definition}
This section describes the synopsis problem more formally than the previous section. Different elements of the problem will be formalized separately. 

\subsection{Problem Inputs}
The two key inputs to the synopsis algorithm are the object tracks represented by tubes and their configuration space.
The configuration space or state space is bound by the $x, y$ and $t$ axes where $x$ and $y$ are the spatial dimensions corresponding to the video frame and $t$ is the temporal dimension corresponding to the duration of the video. For a video, let the frame dimensions be $x_v, y_v$, and total video duration be $t_v$. Since the objects can translate in the 2D plane and also along the temporal axis, we get the configuration space $C = \mathbb{R}^3 $. 

To formalize object tracks, we will first define an object in a video frame. Let there be a set of $n$ objects $A = \{A_1, A_2, ..., A_n\}$ in a video. We can define the object as a set of points that represents a bounded region in the 2D space. Let $O_i$ be the set of points corresponding to the object $A_i$. The shape of all objects will be consistent across all time steps in the video. Therefore, we can define a mapping for every object $f_i: \mathbb \{x_{\mathrm{ic}},y_{\mathrm{ic}}\}\to\mathbb \{x_i,y_i\}$, where $(x_{\mathrm{ic}}, y_{\mathrm{ic}})$ are the centroid co-ordinates of the object $A_i$. Every object can be represented by its centroid and a one-to-many mapping $f$ for every point in the region: \\
\begin{equation}
A_i = \{(x_{\mathrm{ic}},y_{\mathrm{ic}}, f_i(x,y)) \:|\:  \{x,y\} \in O_i \}
\label{eq:object}
\end{equation}

A object track $T_i$ for an object $A_i$ is then defined by:\\
\begin{equation}
T_i = \{ (A_i ,t_i^{\mathrm{start}}, t_i^{\mathrm{end}} )\:|\: 1 \le t_i^{\mathrm{start}} < t_i^{\mathrm{end}} \le t_v\} \label{eq:object_track}
\end{equation}

Here, the objects exists for every time step between $t_i^{\mathrm{start}}$ and $t_i^{\mathrm{end}}$ in the video.

\subsection{Actions}

Actions for this problem include moving the object tracks along the temporal axis. More precisely, it involves the placement of a track at a particular time step.

\subsection{Constraints}
Two constraints are imposed on the algorithm. The first is a time length $t_{\mathrm{max}}$ which imposes a restriction that $t_{\mathrm{last}} \le t_{\mathrm{max}}$, where $t_{\mathrm{last}} = max(t_i^{\mathrm{end}}) \: \forall \: A_i \in A $.

The second is the maximum number of collisions $n_{\mathrm{max}}$ permitted. A collision is when two objects overlap at any time step with an overlap area greater than a fixed threshold $a_{\mathrm{thresh}}$. 
Therefore two objects are said to be colliding if for any time step,

\begin{equation}
    area(O_i \cap O_j) \ge a_{\mathrm{thresh}} \label{eq:collision}
\end{equation}
 
\subsection{Cost Function}

The goal of this algorithm is to minimise the value of $t_{\mathrm{last}}$. The challenge here is the collisions that are incurred in the process of tube placement to minimize $t_{\mathrm{last}}$. Therefore collisions must be penalized and at the same time higher value of $t_{\mathrm{last}}$ must be penalized. Let the synopsis process be defined as $s_u$. The cost is defined as:

\begin{equation}
    E(s_u) = w_0E_c(s_u) + w_1E_t(s_u) \label{eq:cost}
\end{equation}
Here $w_0, w_1$ can be adjusted. $E_c(s_u)$ corresponds to the collision cost and $E_t(s_u)$ corresponds to the temporal cost.

\section{Assumptions}
 Following the above discussions, we can make some assumptions to consider a simplified case for the proposed algorithm.
 \begin{itemize}
     \item The objects will be represented by rectangular bounding boxes. Therefore the $O_i$ will represent a rectangular region.
     \item The bounding boxes cannot be rotated and must stay parallel to the axes. 
     \item Object tracks cannot be shrunk down in any dimension. 
     \item All the geometric representations are available before implementing the synopsis algorithm, i.e., we will consider an offline algorithm.
 \end{itemize}

\section{Proposed method}
\begin{figure}[htbp]
\centerline{\includegraphics[width=0.5\textwidth, height=0.7\textwidth]{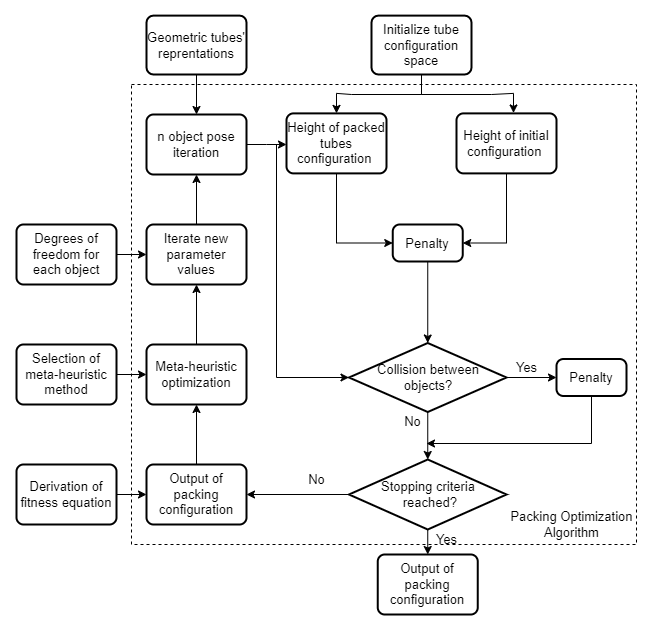}}
\caption{Proposed Methodology derived from \cite{ZHAO2021101234}}
\label{fig9}
\end{figure}

This section will discuss and present a general methodology to solve the problem of video synopsis using packing algorithms. The proposed methodology will be mostly derived from the work by \citet{ZHAO2021101234} on 3D Irregular Object Packing. 3D packing is very similar to a Tetris problem when the nature of the packing problem is online. The methodology proposed in \cite{ZHAO2021101234} includes the selection of a fitness function and an optimization strategy. For video synopsis, the cost function in equation \eqref{eq:cost} will be used as the fitness function. Genetic Algorithm(GA) and Simulated Annealing(SA) are considered the most applicable heuristics for this type of optimization problem \cite{cagan2002survey, oh2018part}. While GA has been used for optimization in video synopsis approaches \cite{gandhi2019object}, we will choose SA as the meta-heuristic solver as experiments done by \cite{ZHAO2021101234} have shown that SA converges to better packing layouts.  The methodology described by \cite{ZHAO2021101234} is modified according to our problem as shown in Fig.~\ref{fig9}. 

In our defined fitness function, $E_c(s_u)$ will be the number of collisions for the current configuration $s_u$, and $E_t(s_u)$ will be the ratio of the height of the packed tubes configuration to the height of the initial configuration. The configuration space will be initialized with the original configuration of the object tracks in the video. The optimization algorithm SA will update the state with random moves in the specified degree of freedom. If the fitness value of the new state is better than the previous state, the new state will be accepted as the current state. Even if the new state's fitness value is worse, it may be accepted with some probability. This ensures that the algorithm does not get stuck in a local minimum.

The key difference between the methodology proposed by \cite{ZHAO2021101234} and ours is that the objects in our case have only one degree of freedom. Additionally, the constraints allow some degree of overlap, which is not allowed in the packing problem.
\section{Applications}
Video synopsis can be applied to every sector where surveillance and monitoring are necessary. Additionally, video synopsis can also be a handy tool for video analytics. Several features can be added on top of the base algorithm such as filtering and trajectory clustering. 

Surveillance and Monitoring applications like warehouse surveillance, security monitoring, and traffic management can be enhanced by filtering the object tracks based on an object characteristic like object color, type, trajectory direction, etc. This reduces the time required for synopsis generation. 

Sports analytics gives us an interesting use case for this problem. We can have discrete frame sequences of a video with the same scene background. Although we do not have a continuous video, we can concatenate the frame sequences or combine the independent object tracks in a separate configuration space. For example, consider a set of videos of the same soccer goalpost. Each video has a different penalty taker. A synopsis generated from this set of videos will provide a unique insight into the penalty shoot patterns. 

Some other interesting applications are video games where the background frame or scene is static. Pac-man is one such example. 

\section{Conclusion}
In this paper, we explored the different data representations of the object tracks as well as the configuration space which are the primary inputs to the synopsis algorithm. The current literature on single camera video synopsis does not include a panoramic camera video synopsis approach. The discussed identifications on the configuration space for these videos can enable the use of existing algorithms for this problem. This paper also presents a problem formulation that is easily transferable to a packing problem. A 3D irregular packing problem methodology is proposed for the problem of video synopsis. Although the 3D packing problem is relatable to the video synopsis problem, there exist differences such as object representation generation, cost function, optimization strategy, and their applications. Lastly, this paper discussed the applications of video synopsis and the variations possible for each of them.


\begingroup
\raggedright
\bibliographystyle{IEEEtranN}
\bibliography{references}
\endgroup

\end{document}